\documentclass[10pt,twocolumn,letterpaper]{article}

\usepackage{iccv}
\usepackage{times}
\usepackage{epsfig}
\usepackage{graphicx}
\usepackage{amsmath}
\usepackage{amssymb}

\usepackage{ragged2e}
\usepackage{booktabs}
\usepackage{color}
\usepackage{multirow}
\usepackage{bm}
\usepackage{bbm}
\usepackage{multirow}
\usepackage{subfigure}
\usepackage{amsmath}

\def\eg{{\em e.g.}}
\def\etal{{\em et al.}}

\newcommand{\figref}[1]{Fig. \ref{#1}}

\newcommand{\br}[1]{\bm{\mathrm{#1}}}
\graphicspath{{figures/},{pics/},{photoes/},{resultpresentation/}}

\usepackage[breaklinks=true,bookmarks=false]{hyperref}

\iccvfinalcopy 

\def\iccvPaperID{****} 

\begin{document}

\title{Image Quality
Assessment for Omnidirectional Cross-reference Stitching}

\author{Jia Li\textsuperscript{1,2}\thanks{Jia Li is the corresponding author. URL: \url{http://cvteam.net}}\qquad
Kaiwen Yu\textsuperscript{1}\qquad
Yu Zhang\textsuperscript{1}\qquad
Yifan Zhao\textsuperscript{1}\qquad
Long Xu\textsuperscript{2}\\
\textsuperscript{1}State Key Laboratory of Virtual Reality Technology and Systems, SCSE, Beihang University\\\
\textsuperscript{2}National Astronomical Observatories, Chinese Academy of Sciences\\
{\tt\small \textsuperscript{1}\{jiali, kevinyu, zhaoyf, octupus\}@buaa.edu.cn \textsuperscript{2}lxu@nao.cas.cn}
}
\maketitle

\begin{abstract}
Along with the development of virtual reality (VR), omnidirectional images play an important role in producing multimedia content with immersive experience. However, despite various existing approaches for omnidirectional image stitching, how to quantitatively assess the quality of stitched images is still insufficiently explored. To address this problem, we establish a novel omnidirectional image
dataset containing stitched images as well as dual-fisheye images captured from standard quarters of 0$^\circ$, 90$^\circ$, 180$^\circ$ and 270$^\circ$. In this manner, when evaluating the quality of an
image stitched from a pair of fisheye images (e.g., 0$^\circ$ and 180$^\circ$), the other pair of fisheye images (e.g., 90$^\circ$ and 270$^\circ$) can be used as the cross-reference to provide ground-truth observations of the stitching regions. Based on this dataset, we further benchmark six widely used stitching models with seven evaluation metrics for IQA. To the best of our knowledge, it is the first dataset that focuses on assessing the stitching quality of omnidirectional images.
\end{abstract}


\section{Introduction}
\begin{figure}[t]
	\begin{center}
		\includegraphics[width=1.0\linewidth]{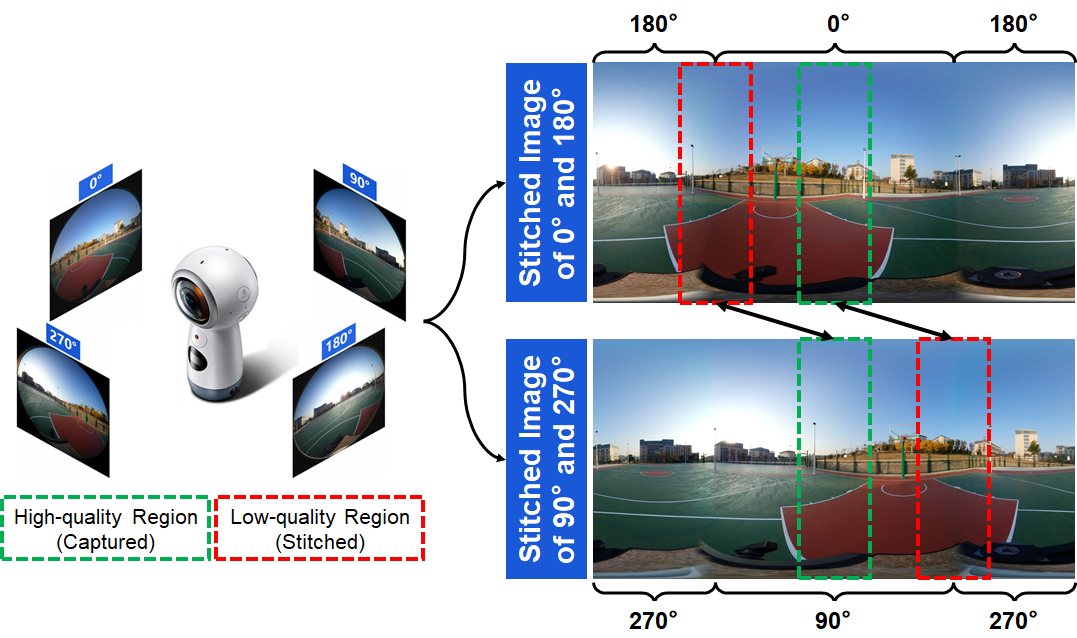}
	\end{center}
	\caption{Motivation of cross-reference stitching quality assessment. We establish a cross-reference omnidirectional dataset with quadruple fisheye images captured at 0, 90, 180, 270 degrees. Taking two images in opposite directions for stitching, the other two images can provide high-quality ground-truth references (green boxes) to assess the stitched low-quality regions (red boxes).}
	\label{fig:motivation}
\end{figure}

With the rapid development of immersive multimedia content in virtual reality (VR), high-quality omnidirectional images are required to provide a natural immersions of real-world scenarios in head-mounted displays. Along with the boost of stitching methods, there exists a huge demand of the automatic quality assessment of stitched omnidirectional images. Actually, the quantitative assessment of stitching quality will be of a great help in the development of VR equipment and computer-aided 3D modeling \cite{baldwin20003d}.

Many image quality assessment (IQA) methods~\cite{DBLP:conf/ncc/NDBCM15, DBLP:conf/cvpr/KangYLD14, DBLP:conf/iccv/LiuWB17, DBLP:conf/iccvw/CheungYTH17, DBLP:journals/tcsv/NiuZGJ18} have been proposed d in the past decades. These IQA researches have gain large successes in common images. Considering the image categories they focuses on, these models can be roughly divided into two categories. The first category mainly focuses on the assessment of common daily images. Along with the denoising and deblurring techniques, many commonly used indexes such as MSE \cite{DBLP:journals/tip/XueZMB14}, PSNR \cite{DBLP:journals/jvcir/Tanchenko14} and SSIM \cite{wang2004image} have been widely used to evaluate the quality of the generated images. Some other studies \cite{bosse2018deep, DBLP:conf/icip/KangYLD15} also use the deep models as the evaluation metric to learn the assessment model automatically. For example, Kang~\etal~\cite{DBLP:conf/icip/KangYLD15} proposed a compact multi-task convolutional neural network for simultaneously estimating image quality and identifying distortions. Liu~\etal~\cite{DBLP:conf/iccv/LiuWB17} further proposed a non-reference image quality assessment with the ranking model of Siamese Network. However, these models usually focuses on the photometric quality indexes such as blurring, noise and color distortions, which may be not suitable for the omnidirectional stitched images.

The second category of the IQA models mainly focuses on the stitched images, which can also be used, more or less, to evaluate the omnidirectional images. In recent years, few studies have explored this less-explored task. For examples, Yang~\etal~\cite{DBLP:conf/iccvw/CheungYTH17} proposed a light-weight model to evaluate the stitched panoramic images based on ghosting and structure inconsistency. However, this proposed metric is designed of normal 2-D plane image stitching, while the dual-fisheye images for generating 360$^\circ$ omnidirectional images face large distortion and information loss in the boundary areas. Duan~\etal~\cite{duan2018perceptual} established an omnidirectional IQA dataset collected with four main distortions of 320 images and proposed a subjective method to assess the image quality. However, this method may have difficulties to get the perfect ground-truth in the stitching areas even with a labour-consuming calibration.

\begin{figure*}[t] 
	\begin{center}
		\includegraphics[width=1.0\linewidth]{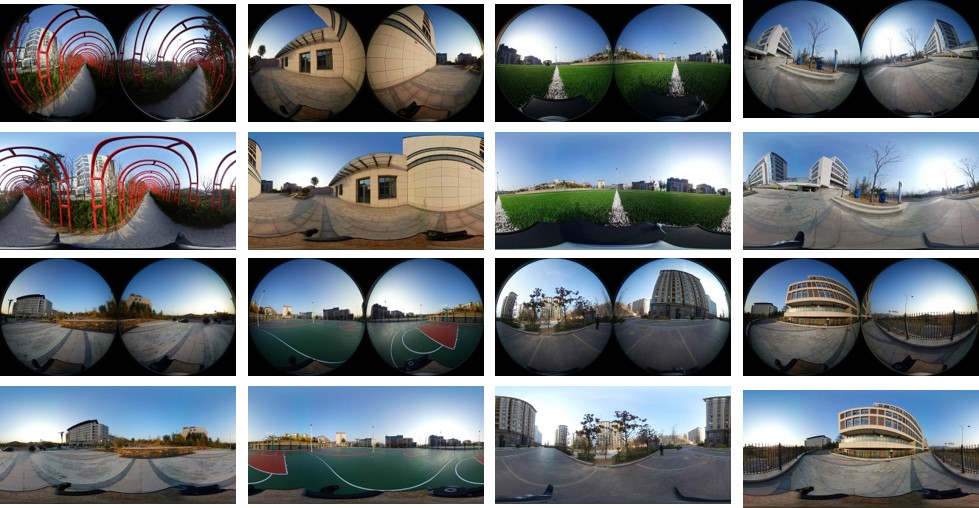}
	\end{center}
	\caption{Outdoor scene images in the proposed CROSS datasets. Collected original fisheye images are shown in the first and third rows, while the ground-truth omnidirectional stitching images are shown in the second and forth rows.}
	\label{fig:outdoor}
\end{figure*}

To tackle these challenges, we mainly address our concerns in the proposed dataset and models. To get the high-quality ground-truth reference for the stitched regions, we resort to the cross-reference dataset establishment. As shown in Fig.~\ref{fig:motivation}, cross-reference is implemented within overlapping region of two wide-angle images for stitching. Taking four fisheye images along a circle, the two images of 0$^\circ$ and 180$^\circ$ are stitched into an omnidirectional image. During stitching, geometry deformation occurs in the region of stitching, so the stitching results of the other two images of 90$^\circ$ and 270$^\circ$ can provide high-quality ground-truth references for measuring such a deformation. To the best of our knowledge, we establish the first cross-reference omnidirectional image dataset which contains 292 quaternions of fisheye images as well as the stitched images generated by seven methods.

\section{Related Work}

Many IQA methods have been proposed in the past decades, which can be roughly grouped into three categories. Some pioneer works for image IQA~\cite{bosse2018deep, DBLP:conf/cvpr/KangYLD14, DBLP:conf/icip/QianLZ13} focuses on both traditional IQA and common panoramic stitching IQA. In this paper, we mainly focuses on the quality assessment omnidirectional images, which is a less-explored task with increasingly demands.

For the quality evaluation of stitched images, the related research results are relatively few. For example, Yang~\etal~\cite{DBLP:conf/iccvw/CheungYTH17} solved the problem of ghosting and structural discontinuity in image stitching by using perceptual geometric error metric and local structure-guide metric, but for immersed image, the evaluation method is not comprehensive enough to detect the global chromatic aberration, and the conditions of blind zone. Huang~\etal~\cite{DBLP:journals/tip/HuangSMBGZC18} proposed the quality evaluation of immersed images, mainly focusing on resolution and compression, neither the quality evaluation of stitching, nor on the image quality evaluation. In \cite{DBLP:conf/icmcs/LingCC18} the authors using convolutional sparse coding and compound feature selection which foucs on the stitching region for stitched image assessment.

There are some existing dataset for panoramic images. For example, Xiao~\etal~\cite{xiao2012recognizing} proposed a SUN360 panorama database covering 360x180 full view for a large variety of environmental scenes, places and the objects within.
However, to the best of our knowledge, the lack of research results on immersed stitching IQA datasets with dual fisheye images, to a certain extent, also restricts the related research development. In the traditional IQA, the existing datasets include LIVE~\cite{DBLP:journals/tip/SheikhSB06} and JPEG2000~\cite{DBLP:journals/tip/MoorthyB11} dataset. While the dataset for stitching image quality evaluation includes SIQA~\cite{DBLP:conf/iccvw/CheungYTH17}, but these mentioned datasets are not suitable for research requirements of 360 $\times$ 180 degree omnidirectional image stitching quality assessment. To this end, we propose a novel OS-IQA database, looking forward to addressing the absence which is convenient to the subsequent study by researchers.

\section{The Cross-reference Dataset}\label{sec:dataset}

\subsection{Omnidirectional Dataset Collection}
To address the absence of omnidirectional stitching IQA dataset, we introduce a novel Cross-Reference Omnidirectional Stitching IQA dataset (CROSS). We use a set of SamSung gear 360 fisheye camera to capture data in various conditions to enhance the robustness.
The proposed dataset is composed of images in 12 various scenarios, which can be further concluded into two families:
\textbf{1) Indoor~:\:} meeting room, class room, stairs, underground park, dance room, lounge, reading-room, lounge
\textbf{2) Outdoor~:\:} street, wild area, basketball court and residential area.

\begin{table}[h]
	\caption{Resolution of images which obtained by several omnidirectional stitching method in detailed.}
	\begin{center}
		\begin{tabular}{cc}
			\hline
			Omnidirectional stitching method & Image resolution \\
			\hline
			SamsungGear & 5472 * 2736 \\
			OpenSource & 2048 * 1024 \\
			WeiMethod & 5418 * 2709 \\
			Stereoscopic Vision Projection & 4290 * 2183 \\
			ManMethod & 5418 * 2709 \\
			Isometric Projection & 5966 * 3091 \\
			Equidistant Projection & 5410 * 2777 \\
			\hline
			\textbf{Fisheye Images} & 5792 * 2896 \\
			\hline
		\end{tabular}
	\end{center}
	\label{tab:images resolution}
\end{table}

\begin{figure*}[t] 
	\begin{center}
		\includegraphics[width=1.0\linewidth]{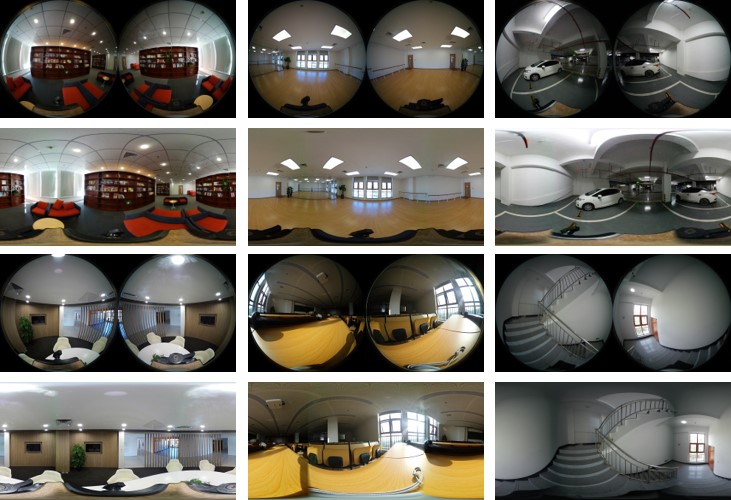}
	\end{center}
	\caption{Indoor scene images in the proposed CROSS datasets. Collected original fisheye images are shown in the first and third rows, while the ground-truth omnidirectional stitching images are shown in the second and forth rows.}
	\label{fig:indoor}
\end{figure*}

Overall, the dataset contains 292 quaternions of fisheye images and the stitching results of seven methods. Some representative examples of outdoor and indoor sceneries can be found in~\figref{fig:outdoor} and~\figref{fig:indoor}. The proposed dataset covers many indoor and outdoor environments as well as natural light and no-natural light conditions.

For the original fisheye images, we take a highest resolution of 5792 $\times$ 2896 limited by the camera settings. For the stitched images, the resolutions are up to the methods, as summarized in Table~\ref{tab:images resolution}. Note that since each scene consists of images captured from various degrees, the synthesis area in each stitching result contains the corresponding ground-truth observations that are required for evaluation.

\begin{figure}[t] 
	\begin{center}
		\includegraphics[width=1.0\linewidth]{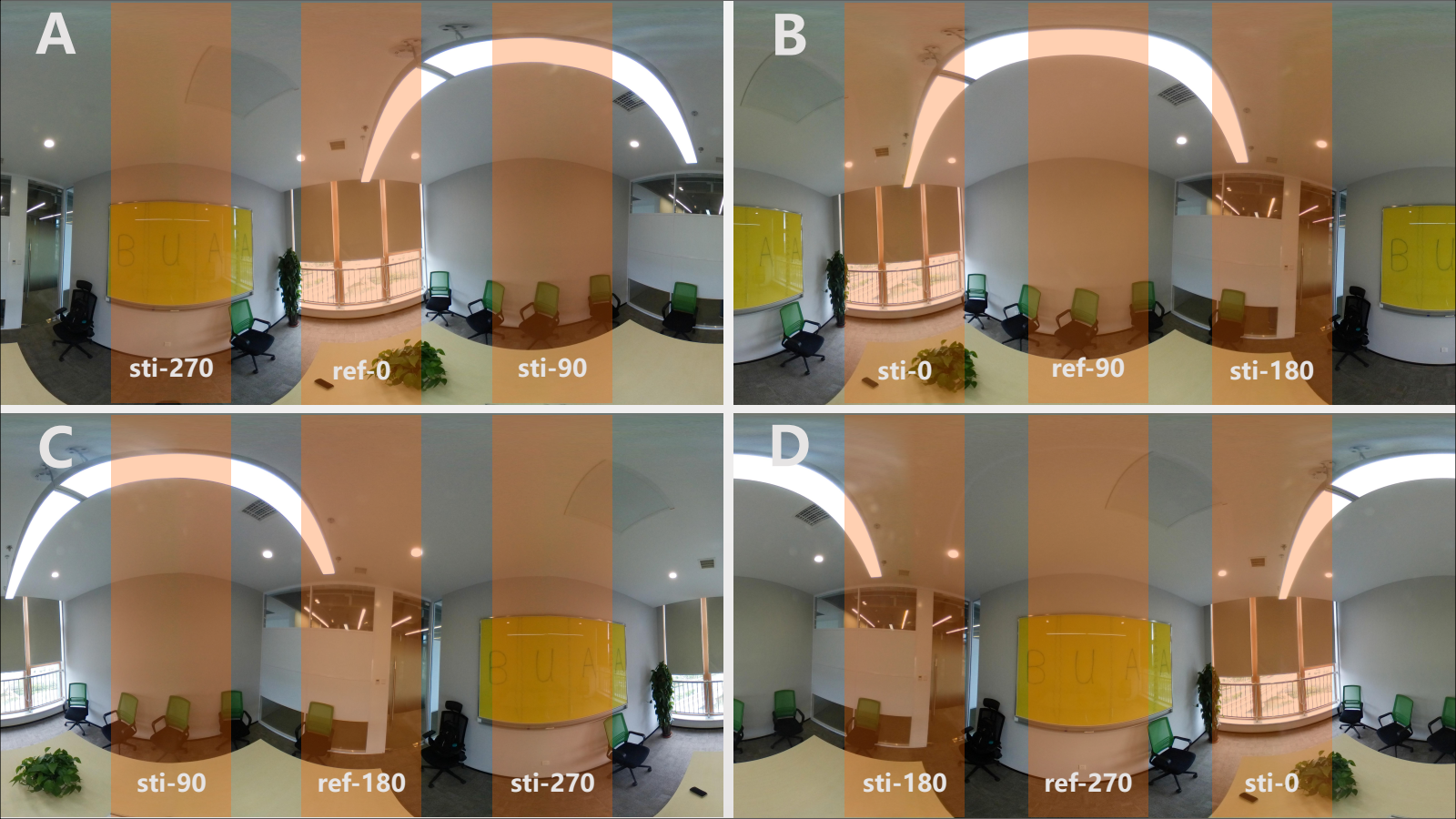}
	\end{center}
	\caption{Cross-reference grouping. One group is consist of four images with different camera angles. The ref-0 is short for reference areas in 0 degrees, and sti denotes the stitching areas. When stitching the images of A and C, the image B and D can provide a perfect ground-truth with less distortions. }
	\label{fig:cross-reference}
\end{figure}

\subsection{Cross-reference Grouping}
To get the highest resolution, we use a set of SamSung gear 360 fisheye camera to capture data in the form of image groups. Each group consists of 4 images captured from different orthogonal categories (0, 90, 180 270) degrees at the same camera position. Taking two images in opposite directions for stitching, there always exists two images can provide ground-truth references without distortions.

To this end, we advocate using the cross-reference images to evaluate the stitching quality. As shown in Fig.~\ref{fig:cross-reference}, the center region of image $\br{B}$ and $\br{D}$ can serve as reference of image $\br{A}$ and $\br{C}$ (and vice versa) due to the orthogonal relationship of degrees. When evaluating the quality of the stitched image at a degree, we call the fisheye images in orthogonal degrees as the cross-reference.

\begin{table*}[t]
	\caption{Average quality assessment scores. The ranking order is viewed in blue.}
	\centering
	\begin{tabular}{ccccccccc}
		\hline
		Method & SamsungGear & OpenSource  & SVP & ManMethod &  IP & EP \\
		\hline\hline 
		MSE~\cite{DBLP:journals/tip/XueZMB14} & 6762.563 & 6287.162 & 7270.892& 5166.097 & 6397.026 & 6279.990 \\
		PSNR~\cite{DBLP:journals/jvcir/Tanchenko14} & 28.050 & 27.878 &  26.204& 29.864 &   27.333& 27.907 \\
		SSIM~\cite{DBLP:journals/tip/WangBSS04}& 0.967 & 0.946 & 0.955 & 1.238 & 1.012& 1.027 \\
		BRISQUE~\cite{DBLP:journals/tip/MittalMB12} & 30.023 & 31.185  & 15.790 & 21.744& 31.669 & 24.732 \\
		NIQE~\cite{DBLP:journals/spl/MittalSB13}& 3.443 & 2.969 & 2.772 & 3.226& 3.230 & 3.306 \\
		PIQE~\cite{DBLP:conf/ncc/NDBCM15}& 32.259 & 45.601 & 23.834 & 29.383 & 30.188& 28.340\\
		CNN~\cite{DBLP:conf/cvpr/KangYLD14}& 21.125 & 19.026 & 19.522 & 18.564 & 20.761 & 19.335 \\
		\hline
	\end{tabular}
	\newline
	\label{tab:specificexperi-data}
\end{table*}

\begin{figure*}[t]
	\centering
	\includegraphics[width=1\linewidth]{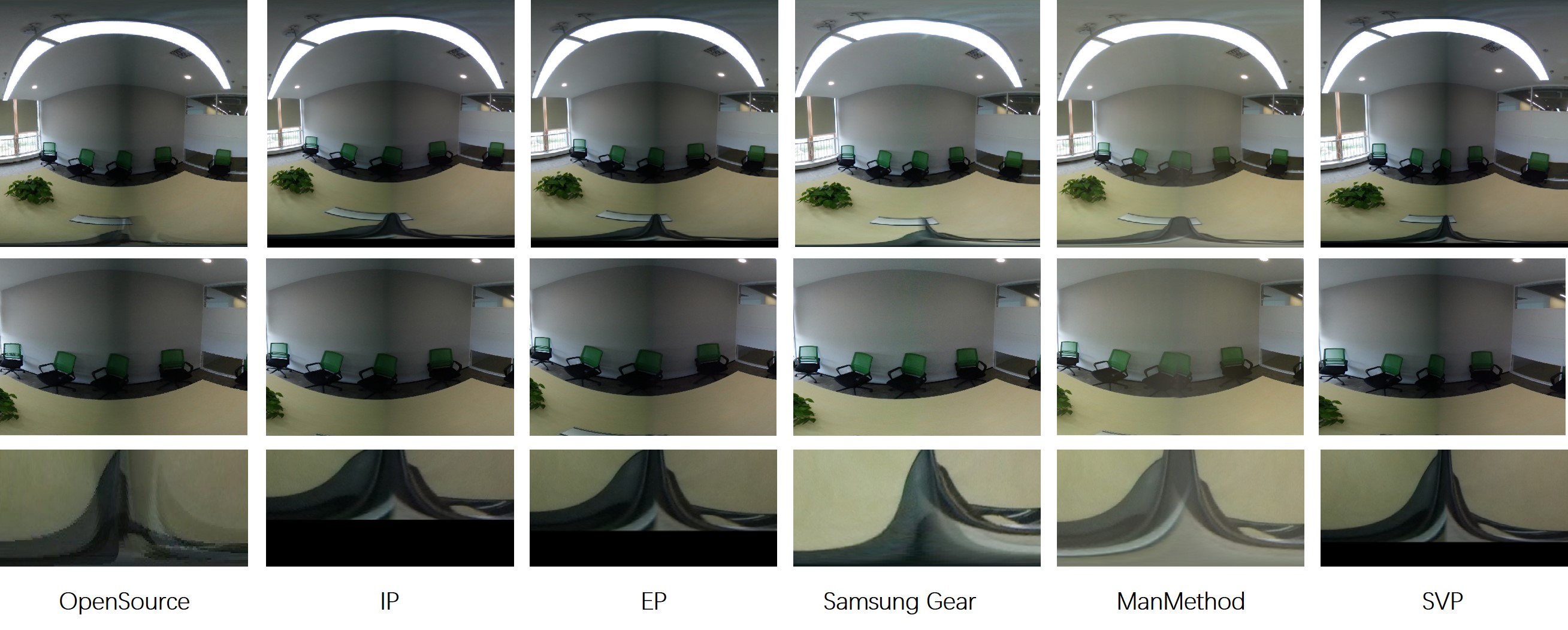}
	\centering
	\caption{Qualitative results of 6 state-of-the-art stitching methods.}
	\label{fig:benchmark}
\end{figure*}

\section{Stitching Quality Benchmark}
\subsection{State-of-the-art Stitching Methods}
To further evaluate the stitching images, we randomly select 192 quaternions from 8 different scenes from our proposed CROSS dataset as the evaluation set. For our learnable human guided classifier, we use the rest 100 dual-fisheye image quaternions as training set.
We adopt 6 widely-used state-of-the-art stitching methods to construct our benchmark, including Samsung Gear 360~\cite{TP-toolbox-web}, OpenSource~\cite{TP-toolbox-web-1}, Stereoscopic Vision Projection (SVP)~\cite{DBLP:conf/grapp/ManeshgarSMP17}, Isometric Projection (IP)~\cite{DBLP:conf/aaai/CaiHH07} and Equidistant Projection (EP)~\cite{DBLP:conf/icmcs/LingCC18}, ManMethod (Manual Method), which finally yields 1344 stitched images in total for comparison.

The EP~\cite{DBLP:conf/icmcs/LingCC18}, IP~\cite{DBLP:conf/aaai/CaiHH07}, and SVP~\cite{DBLP:conf/grapp/ManeshgarSMP17} methods are conducted with different projection strategies which is a unfold-fusion procedure in the spherical coordinates.  Further more, with the development of VR head mount equipment, SamsungGear~\cite{TP-toolbox-web} proposed a model based on the fusion with feature points, and finally adjust the optical parameters (\eg, brightness and chromatism) automatically.

\subsection{Human Subjective Evaluations}
To make a establish of this dataset, we introduce the human subject evaluation as the groundtruth scores to measure these IQA methods.
The human subject evaluation is conducted with 12 expert participants, who evaluated these images with a Head-Mounted Displays. To get the ranking score of these stitching methods, we conduct the pair-wise comparisons of every two images from a same group to get the win-loss times as the final score. In this manner, each group stitched images is compared of $n\times (n-1) /2$ times.
It cost averagely $5\sim7$s for one participants in each comparison.
This win-loss scores for every stitching methods are adopted as Mean Opinion Score (MOS), which represents the human subjective evaluation results.

\subsection{Model Benchmarking}

To evaluate the effectiveness of OS-IQA methods, we compare our methods with 7 widely-used IQA methods, including classical methods
MSE~\cite{DBLP:journals/tip/XueZMB14}, PSNR~\cite{DBLP:journals/jvcir/Tanchenko14}, SSIM~\cite{DBLP:journals/tip/WangBSS04}, no-reference quality assessment metrics, BRISQ~\cite{DBLP:journals/tip/MittalMB12}, NIQE~\cite{DBLP:journals/spl/MittalSB13},
PIQE~\cite{DBLP:conf/ncc/NDBCM15}, and current method based machine learning, CNN-IQA~\cite{DBLP:conf/cvpr/KangYLD14}.

We first use the 7 compared IQA methods to evaluate results of selected stitching models, which finally yields 49 scores, as shown in Table~\ref{tab:specificexperi-data}. To compare with these IQA indexes, we use the ranking order (view in blue) to evaluate these methods. The SamsungGear~\cite{TP-toolbox-web} averagely archives best scores in stitching with minimal distortions and smooth connections while the OpenSource method generate fracture in stitching regions and color distortions.

The qualitative stitching results are shown in~\figref{fig:benchmark}, from which the SamsungGear~\cite{TP-toolbox-web} also generates superior results than the other methods, which is consistent with the quantitative bechmarking results.

\section{Conclusions}
In this paper, we first establish a cross-reference omnidirectional image dataset containing stitched images as well as dual-fisheye images with standard quarters of 0$^{\circ}$, 90$^{\circ}$, 180$^{\circ}$ and 270$^{\circ}$, which will be made publicly available for researchers. In this manner, each quaternion can provide both stitching results and perfect reference groundtruth for quality assessment. Based on this dataset, we further proposed a benchmark to evaluate the state-of-the-art stitching models, From which the Samsung Gear stitching method shows the best performance in image quality. Based on this dataset, we further benchmark 6 widely used stitching models with 7 evaluation IQA metrics. However, this dataset is challenging and would be further a boost to the development of omnidirectional stitching models.
To the best of our knowledge, it is the first attempt that mainly focuses on assessing the stitching quality of omnidirectional images.

{\small
\bibliographystyle{ieee}
\bibliography{IQA}

\begin{thebibliography}{10}\itemsep=-1pt

\bibitem{baldwin20003d}
J.~Baldwin and A.~Basu.
\newblock 3d estimation using panoramic stereo.
\newblock In {\em Proceedings 15th International Conference on Pattern
  Recognition. ICPR-2000}, pages 97--100, 2000.

\bibitem{bosse2018deep}
S.~Bosse, D.~Maniry, K.-R. M{\"u}ller, T.~Wiegand, and W.~Samek.
\newblock Deep neural networks for no-reference and full-reference image
  quality assessment.
\newblock {\em IEEE Transactions on Image Processing}, pages 206--219, 2018.

\bibitem{DBLP:conf/aaai/CaiHH07}
D.~Cai, X.~He, and J.~Han.
\newblock Isometric projection.
\newblock In {\em Proceedings of the Twenty-Second {AAAI} Conference on
  Artificial Intelligence, July 22-26, 2007, Vancouver, British Columbia,
  Canada}, pages 528--533, 2007.

\bibitem{DBLP:conf/iccvw/CheungYTH17}
G.~Cheung, L.~Yang, Z.~Tan, and Z.~Huang.
\newblock A content-aware metric for stitched panoramic image quality
  assessment.
\newblock In {\em 2017 {IEEE} International Conference on Computer Vision
  Workshops, {ICCV} Workshops 2017, Venice, Italy, October 22-29, 2017}, pages
  2487--2494, 2017.

\bibitem{duan2018perceptual}
H.~Duan, G.~Zhai, X.~Min, Y.~Zhu, Y.~Fang, and X.~Yang.
\newblock Perceptual quality assessment of omnidirectional images.
\newblock In {\em 2018 IEEE International Symposium on Circuits and Systems
  (ISCAS)}, pages 1--5, 2018.

\bibitem{TP-toolbox-web-1}
Github.com.
\newblock Dualfisheye.
\newblock \url{https://github.com/ooterness/DualFisheye}, 2016.

\bibitem{TP-toolbox-web}
S.~Group.
\newblock Samsung gear360.
\newblock \url{https://www.samsung.com/global/galaxy/gear-360}, 2017.

\bibitem{DBLP:journals/tip/HuangSMBGZC18}
M.~Huang, Q.~Shen, Z.~Ma, A.~C. Bovik, P.~Gupta, R.~Zhou, and X.~Cao.
\newblock Modeling the perceptual quality of immersive images rendered on head
  mounted displays: Resolution and compression.
\newblock {\em {IEEE} Trans. Image Processing}, pages 6039--6050, 2018.

\bibitem{DBLP:conf/cvpr/KangYLD14}
L.~Kang, P.~Ye, Y.~Li, and D.~S. Doermann.
\newblock Convolutional neural networks for no-reference image quality
  assessment.
\newblock In {\em 2014 {IEEE} Conference on Computer Vision and Pattern
  Recognition, {CVPR} 2014, Columbus, OH, USA, June 23-28, 2014}, pages
  1733--1740, 2014.

\bibitem{DBLP:conf/icip/KangYLD15}
L.~Kang, P.~Ye, Y.~Li, and D.~S. Doermann.
\newblock Simultaneous estimation of image quality and distortion via
  multi-task convolutional neural networks.
\newblock In {\em 2015 {IEEE} International Conference on Image Processing,
  {ICIP} 2015, Quebec City, QC, Canada, September 27-30, 2015}, pages
  2791--2795, 2015.

\bibitem{DBLP:conf/icmcs/LingCC18}
S.~Ling, G.~Cheung, and P.~L. Callet.
\newblock No-reference quality assessment for stitched panoramic images using
  convolutional sparse coding and compound feature selection.
\newblock In {\em 2018 {IEEE} International Conference on Multimedia and Expo,
  {ICME} 2018, San Diego, CA, USA, July 23-27, 2018}, pages 1--6, 2018.

\bibitem{DBLP:conf/iccv/LiuWB17}
X.~Liu, J.~van~de Weijer, and A.~D. Bagdanov.
\newblock Rankiqa: Learning from rankings for no-reference image quality
  assessment.
\newblock In {\em {IEEE} International Conference on Computer Vision, {ICCV}
  2017, Venice, Italy, October 22-29, 2017}, pages 1040--1049, 2017.

\bibitem{DBLP:conf/grapp/ManeshgarSMP17}
B.~Maneshgar, L.~Sujir, S.~P. Mudur, and C.~Poullis.
\newblock A long-range vision system for projection mapping of stereoscopic
  content in outdoor areas.
\newblock In {\em Proceedings of the 12th International Joint Conference on
  Computer Vision, Imaging and Computer Graphics Theory and Applications
  {(VISIGRAPP} 2017) - Volume 1: GRAPP, Porto, Portugal, February 27 - March 1,
  2017.}, pages 290--297, 2017.

\bibitem{DBLP:journals/tip/MittalMB12}
A.~Mittal, A.~K. Moorthy, and A.~C. Bovik.
\newblock No-reference image quality assessment in the spatial domain.
\newblock {\em {IEEE} Trans. Image Processing}, pages 4695--4708, 2012.

\bibitem{DBLP:journals/spl/MittalSB13}
A.~Mittal, R.~Soundararajan, and A.~C. Bovik.
\newblock Making a "completely blind" image quality analyzer.
\newblock {\em {IEEE} Signal Process. Lett.}, pages 209--212, 2013.

\bibitem{DBLP:journals/tip/MoorthyB11}
A.~K. Moorthy and A.~C. Bovik.
\newblock Blind image quality assessment: From natural scene statistics to
  perceptualquality.
\newblock {\em {IEEE} Trans. Image Processing}, pages 3350--3364, 2011.

\bibitem{DBLP:conf/ncc/NDBCM15}
V.~N., P.~D., M.~C. Bh., S.~S. Channappayya, and S.~S. Medasani.
\newblock Blind image quality evaluation using perception based features.
\newblock In {\em Twenty First National Conference on Communications, {NCC}
  2015, Mumbai, India, February 27 - March 1, 2015}, pages 1--6, 2015.

\bibitem{DBLP:journals/tcsv/NiuZGJ18}
Y.~Niu, H.~Zhang, W.~Guo, and R.~Ji.
\newblock Image quality assessment for color correction based on color contrast
  similarity and color value difference.
\newblock {\em {IEEE} Trans. Circuits Syst. Video Techn.}, pages 849--862,
  2018.

\bibitem{DBLP:conf/icip/QianLZ13}
Y.~Qian, D.~Liao, and J.~Zhou.
\newblock Manifold alignment based color transfer for multiview image
  stitching.
\newblock In {\em {IEEE} International Conference on Image Processing, {ICIP}
  2013, Melbourne, Australia, September 15-18, 2013}, pages 1341--1345, 2013.

\bibitem{DBLP:journals/tip/SheikhSB06}
H.~R. Sheikh, M.~F. Sabir, and A.~C. Bovik.
\newblock A statistical evaluation of recent full reference image quality
  assessment algorithms.
\newblock {\em {IEEE} Trans. Image Processing}, pages 3440--3451, 2006.

\bibitem{DBLP:journals/jvcir/Tanchenko14}
A.~Tanchenko.
\newblock Visual-psnr measure of image quality.
\newblock {\em J. Visual Communication and Image Representation}, pages
  874--878, 2014.

\bibitem{DBLP:journals/tip/WangBSS04}
Z.~Wang, A.~C. Bovik, H.~R. Sheikh, and E.~P. Simoncelli.
\newblock Image quality assessment: from error visibility to structural
  similarity.
\newblock {\em {IEEE} Trans. Image Processing}, pages 600--612, 2004.

\bibitem{wang2004image}
Z.~Wang, A.~C. Bovik, H.~R. Sheikh, E.~P. Simoncelli, et~al.
\newblock Image quality assessment: from error visibility to structural
  similarity.
\newblock {\em IEEE transactions on image processing}, pages 600--612, 2004.

\bibitem{xiao2012recognizing}
J.~Xiao, K.~A. Ehinger, A.~Oliva, and A.~Torralba.
\newblock Recognizing scene viewpoint using panoramic place representation.
\newblock In {\em 2012 IEEE Conference on Computer Vision and Pattern
  Recognition}, pages 2695--2702. IEEE, 2012.

\bibitem{DBLP:journals/tip/XueZMB14}
W.~Xue, L.~Zhang, X.~Mou, and A.~C. Bovik.
\newblock Gradient magnitude similarity deviation: {A} highly efficient
  perceptual image quality index.
\newblock {\em {IEEE} Trans. Image Processing}, pages 684--695, 2014.

\end{thebibliography}
}

\end{document}